\documentclass{ws-procs11x85}
\usepackage{ws-procs-thm} 
\usepackage{enumitem}
\usepackage{mathtools}
\DeclarePairedDelimiterX{\infdivx}[2]{(}{)}{%
  #1\;\delimsize\|\;#2%
}

\begin{document}

\title{The Effectiveness of Multitask Learning for \\ Phenotyping with Electronic Health Records Data}

\author{Daisy Yi Ding$^{1}$, Chlo\'e Simpson$^{1}$, Stephen Pfohl$^{1}$, Dave C. Kale$^{2}$, Kenneth Jung$^{1}$, Nigam H. Shah$^{1}$}
\address{$^1$Stanford Center for Biomedical Informatics Research, Stanford University, Stanford, CA; 
\\
$^2$USC Information Sciences Institute, University of Southern California, Marina del Rey, CA}

\begin{abstract}
Electronic phenotyping is the task of ascertaining whether an individual has a medical condition of interest by analyzing their medical record and is foundational in clinical informatics. 
Increasingly, electronic phenotyping is performed via supervised learning. 
We investigate the effectiveness of multitask learning for phenotyping using electronic health records (EHR) data. 
Multitask learning aims to improve model performance on a target task by jointly learning additional auxiliary tasks and has been used in disparate areas of machine learning.  
However, its utility when applied to EHR data has not been established, and prior work suggests that its benefits are inconsistent.  
We present experiments that elucidate when multitask learning with neural nets improves performance for phenotyping using EHR data relative to neural nets trained for a single phenotype and to well-tuned logistic regression baselines.
We find that multitask neural nets consistently outperform single-task neural nets for rare phenotypes but underperform for relatively more common phenotypes.  
The effect size increases as more auxiliary tasks are added.  
Moreover, multitask learning reduces the sensitivity of neural nets to hyperparameter settings for rare phenotypes.
Last, we quantify phenotype complexity and find that neural nets trained with or without multitask learning do not improve on simple baselines unless the phenotypes are sufficiently complex.
\end{abstract}


\copyrightinfo{\copyright\ 2018 The Authors. Open Access chapter published by World Scientific Publishing Company and distributed under the terms of the Creative Commons Attribution Non-Commercial (CC BY-NC) 4.0 License.}


\bodymatter

\section{Introduction}\label{aba:sec1}
The goal of electronic phenotyping is to identify patients with (or without) a specific disease or medical condition using their electronic medical records.  
Identifying sets of such patients (i.e. a patient cohort) is the first step in a wide range of applications such as comparative effectiveness studies \cite{crawford2014emergeing, manion2012leveraging, kaelber2012patient, cholleti2012leveraging}, clinical decision support \cite{longhurst2014green, wei2015extracting}, and translational research \cite{shah2013mining}. 
Increasingly, such phenotyping is done via supervised machine learning methods \cite{agarwal2016learning,Halpern2016b,Aphrodite}.

Multitask learning (MTL) is a widely used technique in machine
learning that seeks to improve performance on a \emph{target task} by jointly modeling the target task and additional
\emph{auxiliary tasks} \cite{caruana1996using}.  
MTL has been used to good effect in a wide variety of domains including computer vision \cite{girshick2014rich,misra2016cross}, natural language processing \cite{plank2016multilingual,liu2017adversarial, collobert2008unified}, speech recognition \cite{arik2017deep,toshniwal2017multitask}, and even drug development \cite{ramsundar2015massively, zhang2014towards}.
However, its effectiveness using EHR data is less well established, with prior work providing contradictory evidence regarding its utility \cite{che2015deep, nori2015simultaneous}.

In this work, we investigate the effectiveness of MTL for phenotyping using EHR data. 
Our preliminary studies on the effectiveness of MTL for phenotyping recapitulated the inconsistent benefits found in prior work \cite{che2015deep,nori2015simultaneous}.  
We thus aimed to elucidate the properties of the phenotypes for which MTL can help versus harm performance.

Our contributions are a systematic exploration of the factors that determine whether or not MTL improves the performance of neural nets for phenotyping compared to neural nets trained for a single phenotype and to well-tuned logistic regression baselines.  
Our experiments suggest the following conclusions: 
\begin{itemize}
  \item MTL helps performance for low prevalence (i.e. rare) phenotypes, but harms performance for relatively high prevalence phenotypes. Consistent with some prior work, there is a dose-response relationship with the number of auxiliary tasks, with the magnitude of the benefit or harm generally increasing as auxiliary tasks are added.

  \item MTL reduces the sensitivity of neural nets to hyperparameter settings. This is of practical importance when one has a limited computational budget for exploring the model space. 

  \item Neural nets trained with or without MTL do not improve on simple baselines unless phenotypes are sufficiently complex. However, learning more complex models can be problematic with complex but low prevalence phenotypes.  We explore this phenomenon by quantifying phenotype complexity using information theoretic metrics.  
    
\end{itemize}

\section{Background}
\subsection{Multitask Networks}

\begin{figure}[t!]
\centering \includegraphics[width=12cm]{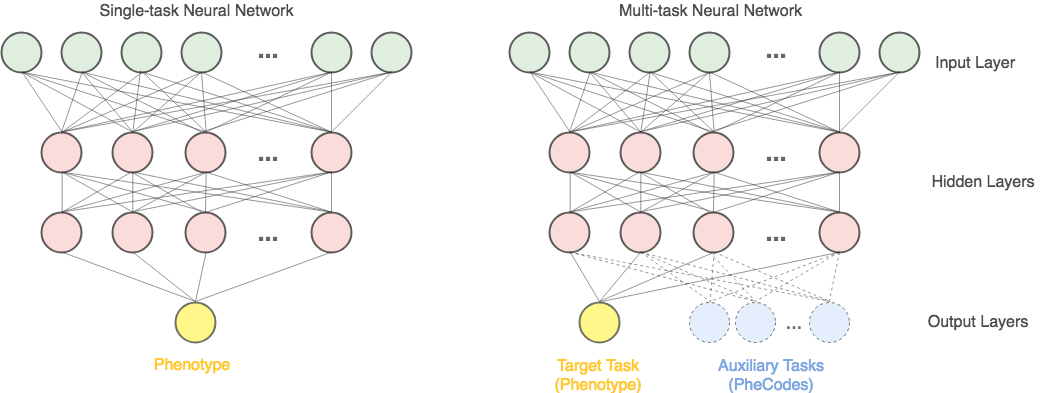}
\caption{The architecture of a multitask neural network for electronic phenotyping is shown on the right: the target task (shown in yellow) and the auxiliary tasks (shown in blue) share hidden layers and have distinct output layers; for comparison, we show the corresponding single-task neural network on the left with a single output layer for the target phenotype.}
\label{method}
\end{figure}

\paragraph{Multitask Learning}
MTL seeks to improve performance on a given target task by jointly
learning additional auxiliary tasks.  
For instance, if the target task is whether or not a patient has type 2 diabetes, one might jointly learn auxiliary tasks such as whether or not the patient has other diseases such as congestive heart failure or emphysema.  
MTL is most frequently embodied as a neural network in which the earliest layers of the network are shared among the target and
auxiliary tasks, with separate outputs for each task (see Figure \ref{method}).  
MTL was originally proposed to improve performance on risk stratification of pneumonia patients by leveraging information in lab values as auxiliary tasks \cite{caruana1996using}.  
It has since been used extensively for health care problems such as predicting illness severity \cite{ghassemi2015multivariate} and mortality \cite{nori2015simultaneous}, and disease risk and progression \cite{ngufor2015multi, wang2014exploring, razavian2016multi, zhou2011multi, lipton2015learning}.  
However, the reported benefits of MTL are inconsistent across problems.  
For example, Che et. al showed that MTL improved performance on identifying physiological markers in clinical time series data \cite{che2015deep}, while Nori et. al concluded that MTL failed to improve performance on predicting mortality in an acute care setting \cite{nori2015simultaneous}.  
Our aim in this study is to clarify when one might expect MTL to help performance on problems using EHR data.  
We focus specifically on the foundational problem of phenotyping, which we discuss next.

\paragraph{Electronic Phenotyping}
In this study, \emph{phenotyping} is simply identifying whether or not a patient has a given disease or disorder.  
The gold standard for phenotyping remains manual chart review by trained clinicians, which is time-consuming and expensive \cite{newton2013validation,overby2013collaborative,mo2015desiderata}.

\begin{figure}[t!]
\centering \includegraphics[width=6.5cm]{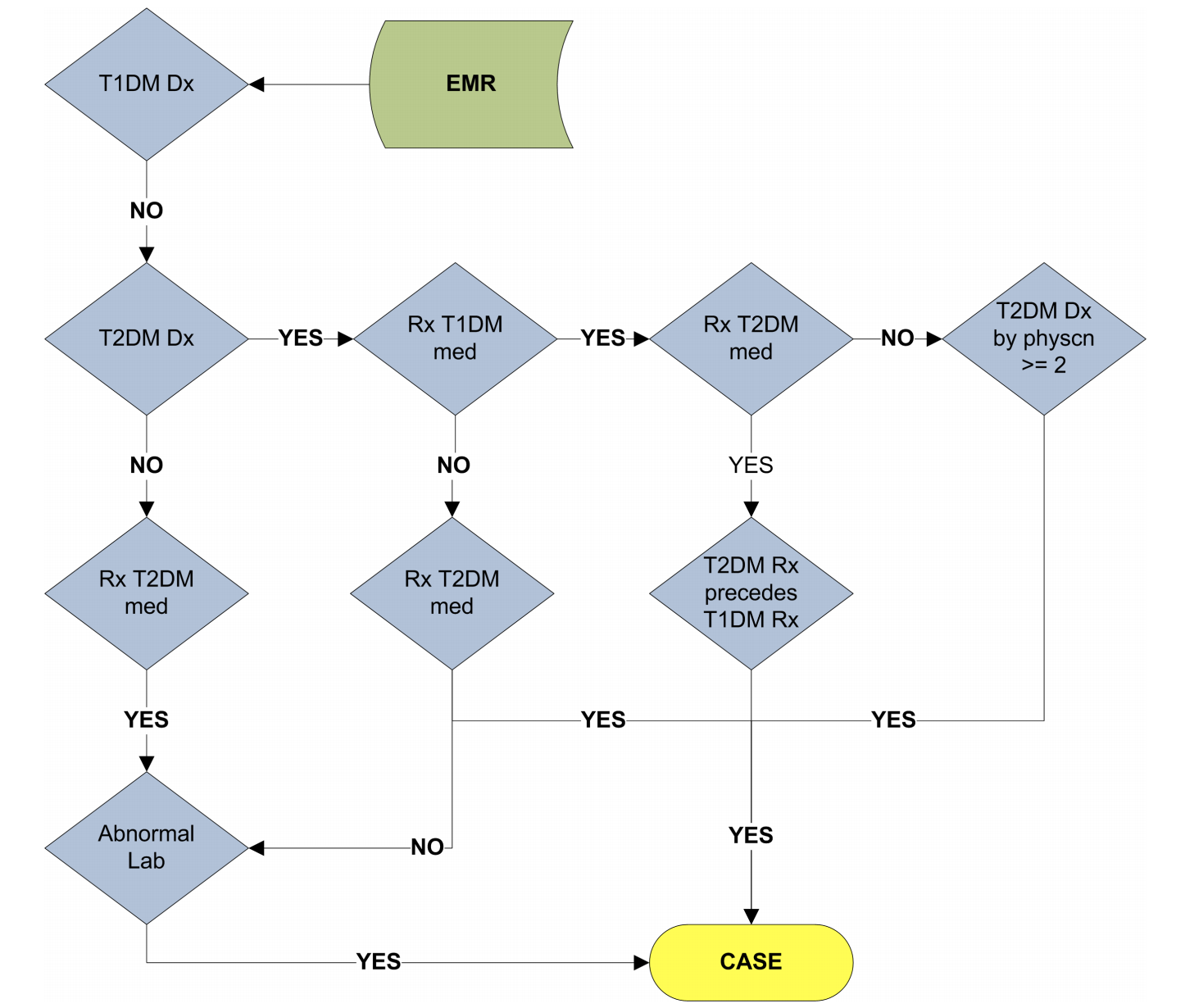}
\caption{Rule-based definitions for \textit{Type 2 Diabetes Mellitus} from PheKB\cite{t2dm}.}
\label{phekb}
\end{figure}

This has spurred work on \emph{electronic phenotyping}, which aims to solve the same problem using automated means and EHR data as input.
The earliest electronic phenotyping algorithms were rule-based
decision criteria created by domain experts \cite{newton2013validation, overby2013collaborative, mo2015desiderata,kho2011electronic, conway2011analyzing}.  
Figure \ref{phekb} shows an example of a rule-based algorithm for type 2 diabetes mellitus.  
In this approach, identifying patients with the phenotype can be automated once the algorithm is specified, but the latter process is still time-consuming and expensive.

More recent work has focused on using statistical learning
\cite{huang2007feature, chen2015building, zhou2014micro, ho2014marble,agarwal2016learning, Halpern2016a} to automate the process of specifying the algorithm itself using the methods of machine learning (i.e. models such as logistic regression,
random forests, and neural nets).  
MTL is a particular method for doing this better.  
Our goal in this work is not to maximize performance for some phenotype but rather to gain insight into when MTL helps versus harms in this approach to phenotyping. 

\section{Methods}
\subsection{Dataset Construction and Design}
\paragraph{Dataset}
Our data comprises de-identified patient data spanning 2010 through 2016 for 1,221,401 patients from the Stanford Translational Research Integrated Database Environment (STRIDE) database \cite{Lowe2009}.
Each patient's data includes timestamped diagnosis (ICD-9), procedure (CPT), drug (RxNorm) codes, along with demographic information (age, gender, race, and ethnicity). 
We use a simple multi-hot feature representation whereby each ICD-9, CPT, and RxNorm code is mapped to a binary indicator variable for whether the code occurs in the patient's
medical history. 
We similarly encode gender, race, ethnicity, and each integer value of age. 
This process results in a sparse representation of 29,102 features.

\paragraph{Target Task Phenotypes}
Phenotyping with statistical classifiers is typically framed as a
binary classification task, which requires data labeled with whether or not the patient has the phenotype.  
For this study, we derive the phenotypes using rule-based definitions from PheKB \cite{kirby2016phekb}, a compendium of phenotype definitions developed to support genome-wide association studies.  
We focus on 4 phenotypes, chosen to span a range of prevalences.  
They are type 2 diabetes mellitus (T2DM), atrial fibrillation (AF), abdominal aneurysm (AA), and angioedema (AE).  
The respective prevalences of these phenotypes in our data are 2.95\%\footnote{The prevalence is low compared to the population prevalence of approximately 9\% because the rule-based definitions from PheKB are tuned for high precision at the cost of lower recall.}, 2.89\%, 0.12\%, and 0.08\%.
We use these rule-based definitions to derive the phenotypes because they are easy to implement, scalable and transparent -- later we describe how we take advantage of the rule-based definitions to gain insight into the effectiveness of MTL relative to baselines.\looseness=-1 

\paragraph{Auxiliary Tasks}
Our auxiliary tasks are to classify \emph{phecodes}, manually
curated groupings of ICD-9 codes originally used to facilitate
phenome-wide association studies \cite{wei2017evaluating}. 
We randomly select phecodes with prevalence between 0.08\% and 2.95\%, i.e. the lowest and the highest target phenotype prevalences, as auxiliary tasks. 
We conduct binary classification on each phecode and experiment with 5, 10, and 20 randomly selected phecodes as auxiliary tasks.

\subsection{Experimental Design}
We aim to investigate whether or not and under what circumstances MTL improves performance upon baselines.  
Recent work suggests that we need to be careful in order to draw robust conclusions on the relative merits of machine learning, especially neural net based methods \cite{Li2017,Melis2017,Lucic2017,Oliver2018}. 

First, one typically randomly partitions data into training,
validation and test sets.  We fit models to the training set, select or tune models using the validation set, and estimate performance on new data using the test set.  
All three steps use finite samples and are thus subject to noise due to sampling.  
This is especially true when data exhibit extreme class imbalance, as is the case with our phenotypes.  
Second, the performance of even simple feed-forward neural nets is known to be sensitive to hyperparameters such as the number of hidden layers and their sizes.  
Finally, fitting neural nets is inherently stochastic due to random initialization of model parameters and training by some variation of stochastic gradient descent.  
This, combined with the highly non-convex nature of neural nets, implies that different training runs of a neural net with fixed
hyperparameters and dataset splits can still result in widely varying performance \cite{keskar2016large}.

We thus designed our experiments to mitigate noise due to these
factors.  
First, for each phenotype, we perform ten random splits of the data into training (80\%), validation (10\%), and test sets (10\%).  
We use stratified sampling to fix the prevalence of the targets to the overall sample prevalence in each of the training, validation and test sets.  
Second, for each of these splits, we perform a grid search over these hyperparameters for the MTNN and STNN models: we vary the number of hidden layers (1 or 2), their size (128, 256, 512, 1024, and 2048), and the initial learning rate for the algorithm (1e-4 and 5e-5). 
Moreover, we performed experiments varying the number of auxiliary tasks (in the form of 5, 10, and 20 nested, randomly selected phecodes) for MTNNs by conducting the above grid search for each scenario. 
For each split, we also fit an L1 regularized logistic regression model, tuned on the validation set.  
We use the area under the Precision-Recall curve (AUPRC) as our evaluation metric since it can be more informative than the area
under the receiver operator characteristic curve (AUROC) in problems with extreme class imbalance \cite{saito2015precision}.  

\paragraph{Phenotype Complexity} \label{complexity_method}
\begin{figure}[t!]
\begin{center}
\includegraphics[width=16cm]{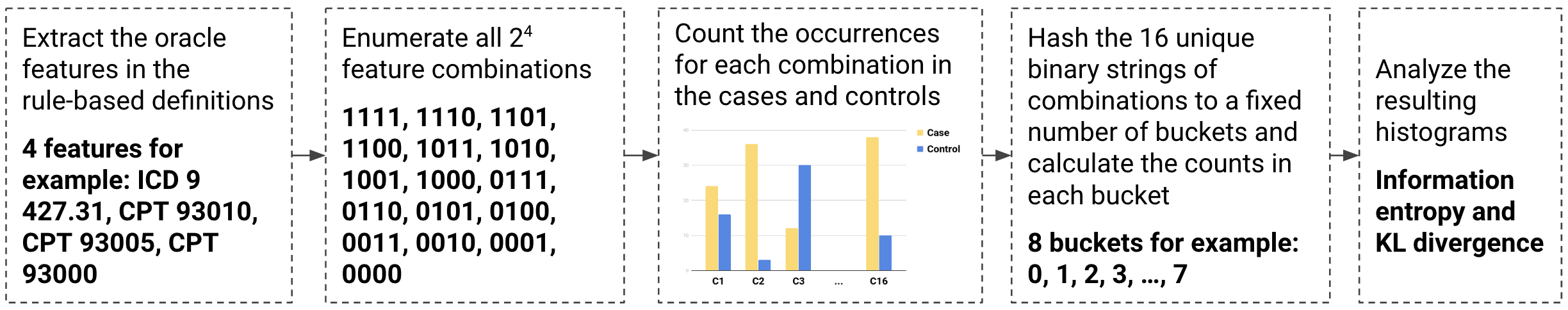}
\end{center}
\caption{An example that illustrates our method on quantifying the phenotype complexity: assume a phenotype has 4 oracle features (for example, ICD 9 427.31, CPT 93010, CPT 93005, and CPT 93000) in its rule-based definitions, we extract the features and enumerate all unique combinations of these features. We represent each combination with a binary string with each digit indicating the presence or absence of an oracle feature. We then count the occurrences of each combination in the positive and negative cases. Afterwards, we use a hash function to map the unique binary strings of combinations to a fewer number of hash codes with each code as a bucket and obtain the counts in each bucket for the positive and negative cases of the phenotype. At last, we analyze the resulting histograms over the buckets with information entropy and KL divergence.}
\label{complexity_method}
\end{figure}

Our experiments suggested that the complexity of the phenotype is important in whether MTNNs and STNNs outperform well-tuned logistic regression. 
We quantified the phenotype complexity with regard to a subset of the features upon which the classifiers are built\footnote{There is no direct way to quantify the complexity of the rule-based definitions shown in Figure \ref{phekb}.}. 
If we had access to an oracle that told us which features of the patient representation are important in determining a patient's phenotype, we could characterize the complexity of the phenotype with regard to the observed combinations of these features in the positive cases.  
We could also compare the distributions of the positive and negative cases to examine how difficult it is to discriminate positive and negative cases given the relevant features.\looseness=-1 

Our phenotypes are derived from the rule-based definitions, which we use as such an oracle: for each phenotype, we extract the features involved in its rule-based definitions (the \emph{oracle features}) and count occurrences of each distinct combination of these features observed in the positive and negative cases.
Each unique combination is represented as a binary string with each digit indicating the presence or absence of an oracle feature.
Since some of the phenotype definitions involve very many combinations, we hash the combinations into a lower-dimensional space, i.e. a fixed number of buckets. 
Specifically, we use a hash function to map the combinations (the variable-length binary strings) to a fixed number of hash codes (the buckets).
We obtain the counts in each bucket for the positive and negative cases and analyze the resulting histograms using two information theoretic metrics.
See Figure \ref{complexity_method} for an example of our method for quantifying phenotype complexity.

Let $\textbf{x}_i$ be the vector of oracle features for bucket $i$.  
We summarize the phenotype complexity of positive cases by treating the histogram as a discrete probability distribution and calculate its information entropy \cite{shannon2001mathematical},
defined as:
\setlength{\abovedisplayskip}{5pt}
\setlength{\belowdisplayskip}{5pt}
\[H(\textbf{X}) = \mathbb{E}_{\textbf{x} \sim P} \left[\log(\textbf{x})\right] = \sum_{i=1}^{n} p(\textbf{x}_i) \log(\textbf{x}_i),\] 
where $n$ is the number of buckets. This metric summarizes the diversity of positive cases with respect to the oracle features and is higher for more complex phenotypes. 

We compare the distributions of the positive and negative cases using the Kullback-Leibler (KL) divergence
\cite{kullback1951information}.  
For discrete probability distributions $P^+$ and $P^-$, the KL divergence from $P^-$ to $P^+$
is defined as:
\[D_{KL}\infdivx{P^+}{P^-} =  \sum_{i=1}^{n} P^+(\textbf{x}_i) \frac{P^-(\textbf{x}_i)}{P^+(\textbf{x}_i)},\] 
where n is the number of buckets\footnote{KL divergence does not admit zero probabilities so we use Laplace smoothing on the distributions to deal with combinations that do not have mutual support.}. $P^+(\textbf{x}_i)$ and $P^-(\textbf{x}_i)$ are the normalized frequencies of bucket $i$ for cases and controls respectively. 
KL divergence measures the dissimilarity between the case and control distributions and is lower for the phenotypes that are harder to discriminate.

\paragraph{Neural Net Details}
All neural nets used ReLU activations \cite{nair2010rectified} for the hidden layers and Xavier initialization \cite{glorot2010understanding} and were trained using Adam \cite{kingma2014adam} with standard parameters ($\beta_{1} = 0.9$ and $\beta_{2} = 0.99$) for 6 epochs\footnote{We found 6 epochs was sufficient for all models to converge.}. 
We controlled overfitting with batch normalization and early stopping on the validation set.  

\section{Experiments and Results}
In this section, we present experiment results that provide insights into the following questions:
\begin{itemize}
\item When does MTL improve performance relative to single-task models for phenotyping?
\item How do the effects of MTL change with the number of auxiliary tasks as defined in the form of phecodes?
\item How do the neural network methods compare with strong
  baseline methods on EHR data, and what are the characteristics of
  the tasks for which they provide some benefit? 
\end{itemize}

\subsection{When Does Multitask Learning Improve Performance?}
\begin{figure}[t!]
\begin{center}
\includegraphics[width=16.5cm]{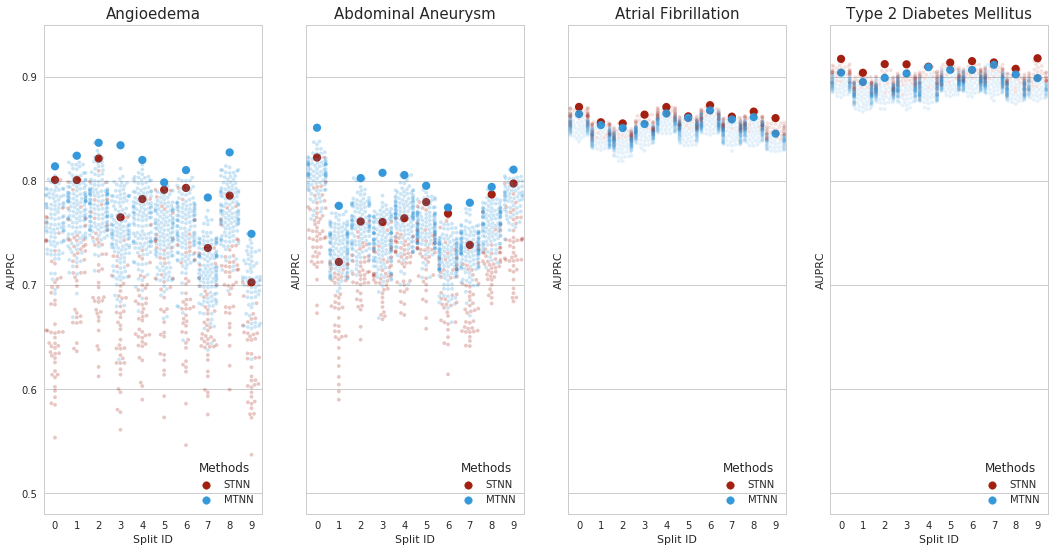}
\end{center}
\caption{MTNN and STNN performance for Angioedema, Abdominal
  Aneurysm, Atrial Fibrillation, and Type 2 Diabetes Mellitus with various hyperparameter settings across the ten splits; the best case MTNN and STNN performance is emphasized by solid dots: the blue dots correspond to MTNNs and the red dots correspond to STNNs.}
\label{result_1}
\end{figure}

\begin{figure}[t!]
\centering
\begin{center}
\includegraphics[width=12cm]{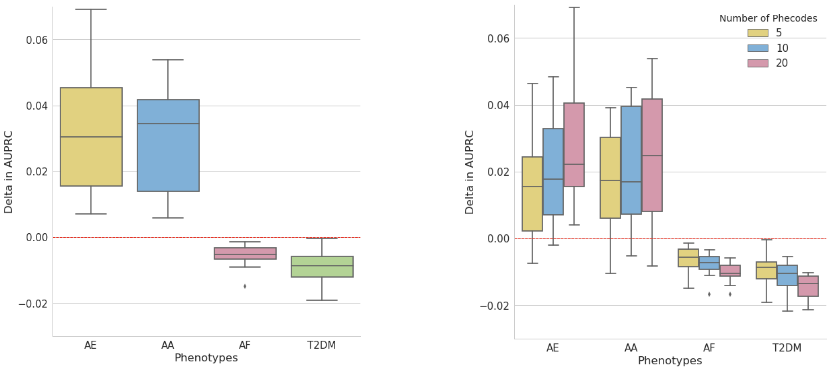}
\end{center}
\caption{The left plot shows the pairwise differences in AUPRC values of the optimal MTNNs and STNNs for Angioedema, Abdominal Aneurysm, Atrial Fibrillation, and Type 2 Diabetes Mellitus across the ten splits. The right plot shows the pairwise differences in AUPRC values of the optimal STNNs and MTNNs with different number of phecodes as auxiliary tasks.}
\label{result_2}
\end{figure}

We investigate the performance of MTNNs over a range of hyperparameter settings and over multiple random splits of the data.  
MTNNs performance is compared to the performance of STNNs over the same hyperparameter settings and data splits. 
Figure \ref{result_1} shows the optimal MTNN and STNN performance achieved on each split for the four phenotypes. 
We find that MTNNs consistently outperform STNNs for the low prevalence phenotypes, i.e. angioedema and abdominal aneurysm.  
In contrast, MTL harms performance for the relatively high-prevalence phenotypes, i.e. T2DM and atrial
fibrillation. 
The left plot in Figure \ref{result_2} shows the pairwise differences between MTNN and STNN optimal performance across the splits. 

Moreover, the performance of STNNs is very sensitive to hyperparameter settings for the low prevalence phenotypes, as illustrated by the large spread in AUPRC values (see Figure
\ref{result_1}). 
In contrast, MTNNs are more robust to hyperparameter settings for these phenotypes. 
In practice, tuning neural nets is time-consuming and finding an ideal model demands extensive computation. 
MTL may increase our chance of finding a reasonable model, which is of practical value when one has a limited computational budget on model space exploration. 

\subsection{Relationship Between Performance and Number of Tasks}
We investigate how MTL is influenced by the number of auxiliary tasks (defined in the form of phecodes). 
We trained MTNNs with nested sets of 5, 10, and 20 randomly selected phecodes (i.e. the 5-phecode set is a subset of the 10-phecode set, and so on), and reported the performance with the optimal hyperparameter setting for each split. 
The right plot in Figure \ref{result_2} shows pairwise differences in AUPRC values between MTNNs and STNNs. For the low prevalence phenotypes, more phecodes increases performance gains.  
Similarly, more phecodes for high prevalence phenotypes leads to more severe negative effects, though the scale of the negative effects is smaller than the positive effects for low prevalence phenotypes\footnote{This dose-response relationship with the number of auxiliary tasks recapitulates the findings of Ramsundar et al \cite{ramsundar2015massively}, but we find the relationship holds for both the benefit and harm of MTL.}.  

\subsection{Comparison with Logistic Regression Baseline}
\begin{figure}[t!]
\begin{center}
\includegraphics[width=16.5cm]{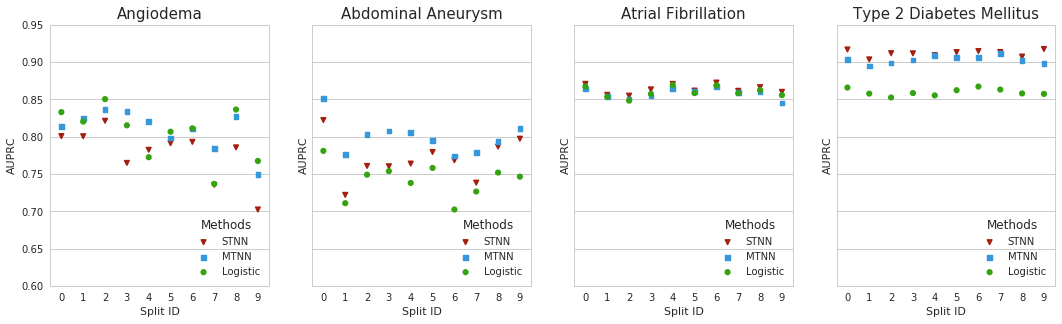}
\end{center}
\caption{MTNN, STNN, and LR optimal performance for Angioedema, Abdominal Aneurysm, Atrial Fibrillation, and Type 2 Diabetes Mellitus across the ten splits: the blue squares correspond to MTNNs, the red triangle correspond to STNNs, and the green dots correspond to L1 regularized logistic regression.}
\label{result_5}
\end{figure}

In discussing the merits of MTL, it is important to also compare the performance against simpler baseline methods in addition to single-task neural nets.  
We compare the performance of the neural nets with L1 regularized
logistic regression (LR), a consistently strong baseline for EHR data \cite{rajkomar2018scalable,razavian2015population} (see Figure \ref{result_5}). 
LR is consistently outperformed by the neural nets for abdominal
aneurysm and type 2 diabetes mellitus, which are low and high prevalence respectively. 
For angioedema, a low prevalence phenotype, performance relative to LR is inconsistent across the splits, although MTNNs consistently beat STNNs. And for atrial fibrillation, a high prevalence phenotype, MTNNs and STNNs provide little or no benefit over LR.  
Prevalence alone is insufficient to account for the relative performance between both MTNN and STNN and LR. 

\subsection{Interaction between Phenotype Prevalence and Complexity}
Our comparison of MTNNs and STNNs versus LR suggests that phenotype prevalence alone cannot explain when neural nets outperform simpler linear models.  
We hypothesized that phenotype complexity also plays a role since neural nets with or without MTL can automatically model non-linearities and interactions, while LR must have non-linearities and interactions explicitly encoded in features.  
We leveraged the rule-based phenotype definitions to explore this hypothesis and found evidence of an interaction between phenotype prevalence and complexity.  

\begin{figure}[t!]
\begin{center}
\includegraphics[width=16cm]{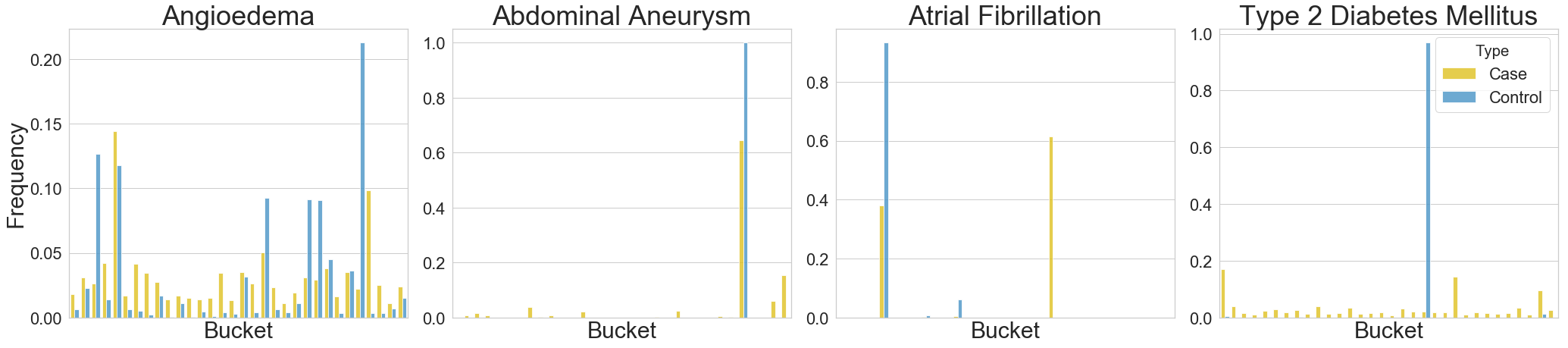}
\end{center}
\caption{Distributions of the combinations of the oracle features involved in the rule-based definitions for Angioedema, Abdominal Aneurysm, Atrial Fibrillation, and Type 2 Diabetes Mellitus. The yellow and blue bars correspond to the positive and negative cases respectively. The x-axes represent the buckets of unique combinations of the oracle features: in our study, we use 32 buckets. Note that the choice of 32 buckets was arbitrary and not tuned in any way.}
\label{complexity}
\end{figure}

\paragraph{Phenotype Complexity}
For each phenotype, we generated histograms of the observed combinations of the oracle features for the positive and negative cases (see Figure \ref{complexity}) and calculated the information entropy of the positive cases and the KL divergence between the positive and negative cases (see Table \ref{entropy and kl}) as described in Methods \ref{complexity_method}.  

\begin{table*}[t!]    
\tbl{Phenotype Complexity}
    {\begin{tabular}{@{}lccc@{}}\Hline Phenotype & Prevalence &
        Entropy & KL Divergence\\ 
        \Hline 
        Angioedema & 0.08 \% & 3.233 & 0.930 \\ 
        Abdominal Aneurysm & 0.12\% & 1.396 & 2.414 \\
        Atrial Fibrillation & 2.89\% & 0.709 & 5.383 \\ 
        Type 2 Diabetes Mellitus & 2.95 \% & 3.012 & 3.806 \\ 
        \Hline
\end{tabular}}
\label{entropy and kl}
\end{table*}

We find that atrial fibrillation, a high-prevalence phenotype, has low entropy and high KL divergence.  
With respect to the oracle features, all the positive cases are similar to each other, while the positive and negative cases are very dissimilar to each other.
A relatively simple model should be able to capture this, explaining the observation that LR achieves comparable performance to MTNNs and STNNs for this phenotype. 

Abdominal aneurysm, a low prevalence phenotype, and T2DM, a high prevalence phenotype, have higher information entropy and lower KL divergence values than atrial fibrillation.  
Thus, the positive cases are more diverse and discrimination is more difficult than atrial fibrillation with respect to each phenotype's oracle features.  
For these phenotypes, both MTNNs and STNNs outperform LR -- we benefit from more expressive models. 
However, whether MTNNs beat STNNs depends on prevalence.  

Finally, angioedema has the highest entropy and lowest KL divergence -- it is both the most complex and hardest to discriminate of the four phenotypes.  
Complex phenotypes should benefit from more expressive models. 
However, we observe that while MTNNs consistently outperform STNNs, their performance relative to LR is inconsistent across splits.  
One possible explanation for this behavior is that relative performance is sensitive to the assignment of patients to training, validation and test sets: with such diverse cases and common support with respect to the oracle features, it is much more likely for the test set to contain patients unlike any seen in the training set.  

\section{Limitations}
We have set out to investigate MTL and its effectiveness for electronic phenotyping.  
However, our work has important limitations.  
First, we randomly select phecodes for auxiliary tasks, but it has been argued that auxiliary tasks should be directly related to the target task \cite{caruana1997multitask}. 
It is possible that better auxiliary tasks would improve the benefit of MTL. 
Specifically, more related phecodes might mitigate or eliminate the performance degradation observed for the high-prevalence phenotypes or inconsistent relative performance between MTNN and LR for angioedema.
However, the notion of task relatedness is underspecified so it is problematic to compute in order to select auxiliary tasks.  
Indeed, in preliminary work, we explored various formulations of relatedness to select auxiliary tasks but found that none performed better than random selection.  
One could ask domain experts to manually construct or pick auxiliary tasks for specific phenotypes, but this is beyond the scope of this work.
Moreover, it has also been shown that the task relatedness is unnecessary for MTL to provide benefits \cite{romera2012exploiting}. 
However, we acknowledge that it is an interesting line of inquiry for future work to further explore how to improve multitask learning for electronic phenotyping.
Second, to address the unavailability of large-scale ground truth phenotypes, we use rule-based definitions because they are transparent and available, but we recognize that the phenomenon we observe may be artifacts of the rule-based definitions. 
We also acknowledge the possibility that the observed phenomenon might not generalize to other phenotypes; we focused on four phenotypes to conduct an in-depth examination, sacrificing breadth.
Finally, the rule-based phenotype definitions contain predicates encoding temporal relationships, e.g., a drug code followed by a diagnosis code. 
Our simple multi-hot feature representation does not encode temporal information.
As a result, there is an upper bound on the performance of any statistical classifier using this feature representation.  

\section{Conclusion}
We have investigated the effectiveness of multitask learning on electronic phenotyping with EHR data, aiming to elucidate the properties of situations for which MTL improves or harms performance. 
We trained multitask neural networks to classify a target phenotype jointly with auxiliary tasks drawn from phecodes. 
We found that MTL provided consistent performance improvements over single-task neural networks on extremely rare phenotypes.  
However, for relatively higher prevalence phenotypes, MTL actually reduced performance.  
In both cases, the effect scaled with the number of auxiliary tasks as defined in the form of phecodes. 
Moreover, we found that MTL improved the robustness of neural networks to hyperparameter settings for the extremely rare phenotypes, which is of practical value in situations when one has a limited computational budget for model exploration. 
Finally, we analyzed phenotype complexity to shed light on the relative performance of both MTNN and STNN versus well-tuned L1 regularized logistic regression baselines and found evidence of an interaction between phenotype prevalence and complexity. 
We showed that simple linear models are sufficient for non-complex phenotyping tasks.  
More expressive models can substantially improve performance for more complex phenotypes, but only if the data support learning them well, which may be problematic for rare phenotypes.  

\section*{Acknowledgments}
This work was supported by NLM R01-LM011369-05 and a grant supporting the Observational Health Data Science and Informatics (OHDSI) by Janssen Research and Development LLC. Internal funding by the School of Medicine at Stanford also supported part of this work.  We gratefully acknowledge Jason Fries for many helpful discussions about this work.  

\bibliographystyle{ws-procs11x85} 
\bibliography{ws-procs11x85}

\end{document}